\documentclass{article}
\usepackage{ijcai23}

\usepackage{times}
\usepackage{soul}
\usepackage{url}
\usepackage[hidelinks]{hyperref}
\usepackage[utf8]{inputenc}
\usepackage[small]{caption}
\usepackage{graphicx}
\usepackage{amsmath}
\usepackage{amsthm}
\usepackage{booktabs}
\usepackage{algorithm}
\usepackage{algpseudocode}
\usepackage[switch]{lineno}
\usepackage{xcolor}
\usepackage{multirow}
\usepackage{makecell}
\newtheorem{definition}{Definition}
\title{FedBone: Towards Large-Scale Federated Multi-Task Learning}

\author{
Yiqiang Chen$^{1, 2}$
\and
Teng Zhang$^{1, 2}$\and
Xinlong Jiang$^{1, 2}$\and
Qian Chen$^{1, 2}$\and
Chenlong Gao$^{1, 2}$ \\\And
Wuliang Huang$^{1, 2}$
\affiliations
$^1$Beijing Key Lab. of Mobile Computing and Pervasive Devices, Inst. of Comp. Tech., CAS\\
$^2$University of Chinese Academy of Sciences, Beijing, China\\
\emails
\{yqchen, zhangteng19s, jiangxinlong, chenqian20b, gaochenlong, huangwuliang19b\}@ict.ac.cn
}

\begin{document}

\maketitle

\begin{abstract}
Heterogeneous federated multi-task learning (HFMTL) is a federated learning technique that combines heterogeneous tasks of different clients to achieve more accurate, comprehensive predictions.
%在现实应用中的视觉与自然语言任务通常需要大规模模型来提取高级抽象特征
In real-world applications, visual and natural language tasks typically require large-scale models to extract high-level abstract features.
However, large-scale models cannot be directly applied to existing federated multi-task learning methods.
Existing HFML methods also disregard the impact of gradient conflicts on multi-task optimization during the federated aggregation process.
% 在联邦聚合过程中忽略了梯度冲突对多任务优化的影响
% 通过服务器端的大规模通用特征提取器，并在客户端上进行本地任务适配能够平衡多任务的泛化性和特定任务的个性化。
% 聚合各客户端任务，在多任务的泛化性和任务特定的个性化上，都有重要意义。
% 现有的方法只做了数据异构性，没有关注到任务的异构性
% 因此我们提出了
% Integrating Federated Learning (FL) to deploy large-scale pre-trained models have gained increasing attention due to their practical value in various fields, including medical healthcare where data privacy is a top priority. However, clients often lack the computational resources to run full-size models, and their tasks are often heterogeneous, making traditional FL unsuitable.
% 从服务端-客户端分离学习和梯度映射两个视角实现了泛化性更好的大规模模型
In this work, we propose an innovative framework called $\mathtt{FedBone}$, which enables the construction of large-scale models with better generalization from the perspective of server-client split learning and gradient projection. 
% 我们将要联邦学习的模型被分成了两个部分
We split the entire model into two components: a large-scale general model (referred to as \emph{the general model}) on the cloud server and multiple task-specific models (referred to as \emph{the client model}) on edge clients, solving the problem of insufficient computing power on edge clients.
The conflicting gradient projection technique is used to enhance the generalization of the large-scale general model between different tasks.
%The proposed framework is evaluated on two benchmark datasets and a real ophthalmic dataset. Comprehensive results demonstrate that $\mathtt{FedBone}$ efficiently adapts to heterogeneous local tasks of each client and outperforms existing federated learning algorithms in most dense prediction and classification tasks with less than $1/10$ computational resources required on the client side.
The proposed framework is evaluated on two benchmark datasets and a real ophthalmic dataset. Comprehensive results demonstrate that $\mathtt{FedBone}$ efficiently adapts to heterogeneous local tasks of each client and outperforms existing federated learning algorithms in most dense prediction and classification tasks with off-the-shelf computational resources on the client side.
\end{abstract}

\section{Introduction}
%%%%%%%%%%%%%%%%%%%%%%%%%%%都是GPT写的，需要改%%%%%%%%%%%%%%%%%%%%%%
%%%%%% 大规模预训练模型推动了NLP和CV等领域的发展，并广泛应用于多种场景，大模型需要大算力和大数据，但在一些场景下并不能满足。比如在医疗场景下，很多医院本身并没有大算力，而且交隐私敏感的医疗数据交给高算力机构去训练大模型也是不合规不安全的 %%%%%%
%Large-scale pre-trained models have exerted a significant impact on the advancement of the natural language processing (NLP) and computer vision (CV) fields. These models have found broad applicability in diverse scenarios, including image recognition, machine translation, and sentiment analysis. Nonetheless, the growing complexity of these models can result in exceedingly high computational demands, particularly in contexts where large-scale computing power and data storage are unavailable. One scenario where this constraint is especially pronounced is the healthcare industry, where hospitals and clinics store vast volumes of sensitive data. Concerns over privacy and security make it difficult to entrust third-party institutions with the task of training models on medical data that are sensitive. Moreover, many hospitals may lack the resources to purchase or lease computing resources needed to train such scale models.
% 2023.5.11 Qian Chen
The progress of Natural Language Processing (NLP) and Computer Vision (CV) has been significantly driven by the evolution of large-scale pre-trained models. Nonetheless, the efficacy of these extensive models predominantly hinges on substantial computational resources and extensive datasets. This presents challenges in situations where such resources are not readily accessible or practically feasible to employ.
%One particular domain facing these challenges is healthcare, where numerous medical institutions may lack the computational power required for training and implementing large-scale models. 
Furthermore, the sensitivity and privacy concerns associated with medical data render it impractical and insecure to entrust such data to external high-computational organizations for training large models.
% These limitations necessitate alternative approaches that can accommodate the unique requirements of the healthcare industry while ensuring compliance with privacy regulations.
In this context, the concept of Federated Learning~\cite{mcmahan_communication-efficient_2017} emerges as a promising solution. Federated learning enables collaborative model training across distributed edge devices or institutions while keeping the data local and confidential. By leveraging the potential of distributed learning, federated learning enables institutions to participate in model training without compromising data privacy or necessitating significant computational resources.

\begin{figure}[t]
\includegraphics[width=\linewidth]{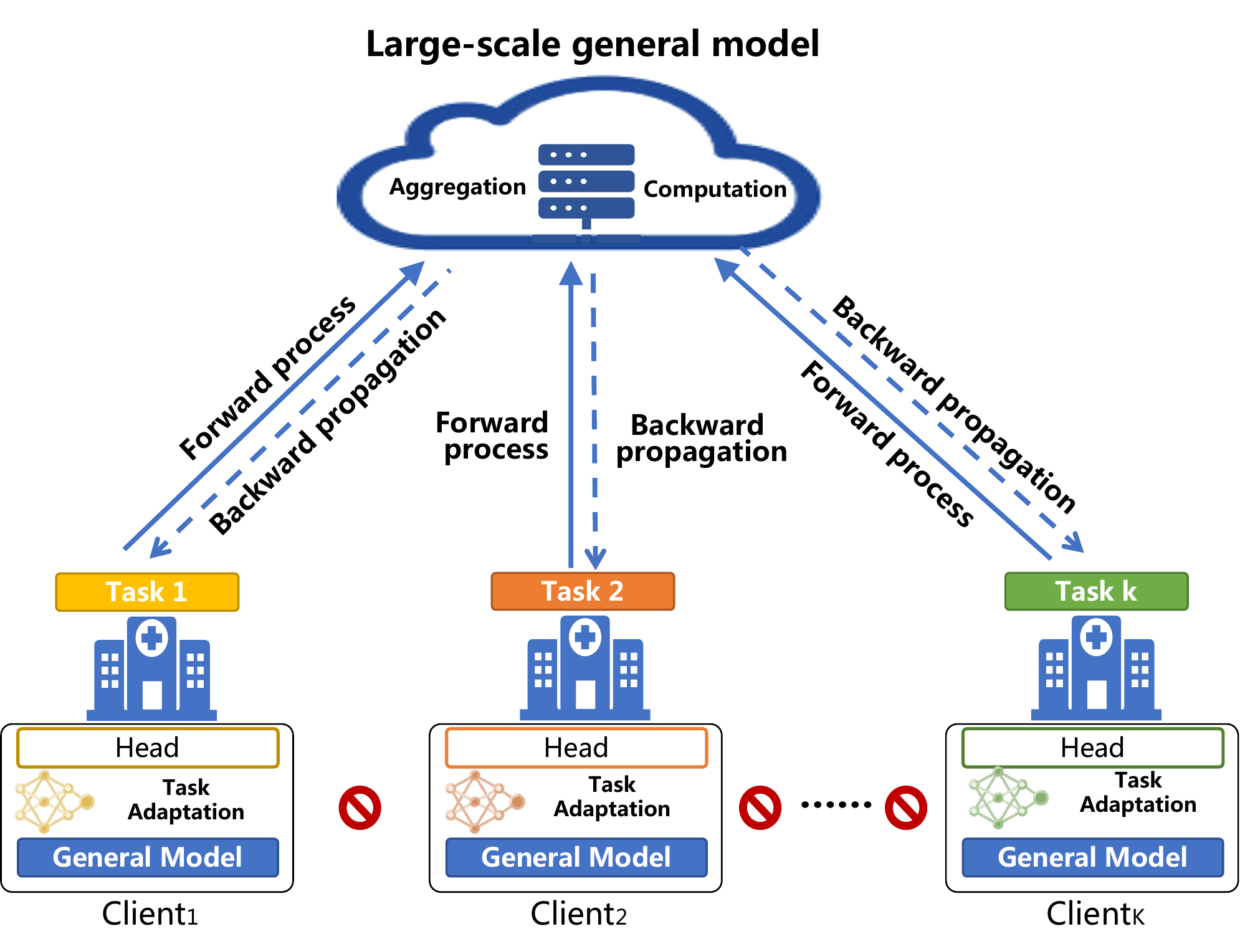}
\caption{The overview of our proposed framework}
\label{fig:small_overview}
\end{figure}

Training large-scale models using federated learning faces two primary challenges. Firstly, there is the issue of resource constraints at the edge. Many edge devices, such as edge computing machines and Internet of Things (IoT) devices, have limited computational capabilities, making it difficult to train and deploy large models directly on these devices. Secondly, optimizing multiple heterogeneous tasks among participating parties becomes challenging especially in a federated setting. % Different applications require distinct processing of the same modality of data, resulting in diverse downstream tasks. 
% Large-scale models are appealing because they can be applied to these heterogeneous tasks, requiring only fine-tuning the pre-trained model for every task to achieve better performance than individual training. However, optimizing multiple heterogeneous tasks among participating parties becomes challenging especially in a federated setting. 
Existing personalized federated and multi-task federated methods~\cite{cao_cross-silo_2023,shu_clustered_2023} have mainly focused on addressing the heterogeneity of data distributions but have disregarded the consideration of task heterogeneity. The heterogeneity of tasks results in different optimization objectives for each task, and simply aggregating models trained on diverse tasks can lead the federated model to optimize in a biased direction, which results in a decrease in the generalization ability of federated large-scale models, consequently diminishing their practical utility.

In light of these challenges, we propose $\mathtt{FedBone}$, a federated multi-task learning framework as shown in  Figure \ref{fig:small_overview}, which takes advantage of the server-client split learning paradigm to enable the edge clients to participate in large-scale federated training with low memory footprints.
The $\mathtt{FedBone}$ framework is designed to execute a multi-stage process for handling heterogeneous client tasks which entail client-side data embedding, server-side universal model feature extraction, and client-side task-specific processing. Throughout the process, edge clients are only responsible for computing data embedding and propagating it to the cloud server for feature extraction of the large-scale general model. The resulting latent representations are then dispatched back to the clients to perform task output.
% The process of the heterogeneous client tasks involves local data embedding, server-side feature extraction of the universal model, and local task output.
To enhance the general model's generalization, we introduce a gradient projection method and gradient rescaling based on historical gradient attention to reduce the negative impact of conflicting gradient components on gradient aggregation.
% In federated multi-task learning, it's possible to view each client as a task. Hence, when describing the aggregation of the general model, we'll use interchangeably "task" and "client", "task gradient" and "client gradient".
The task output module on the client is tailored to specific task types but is generally concise due to the assumption of feature extraction having already been fulfilled. Latent representations extracted from the large-scale general model are usually low-level features for various tasks. %, rather than task-specific high-level semantic features.
Therefore, we propose a task adaptation module, which utilizes deformable convolutions and a self-attention mechanism to focus on low-level features in the task-specific region and perform task interactions.
%The proposed task adaptation module significantly improves downstream task performance in the experiments.
Our main contributions can be summarized as follows:
\begin{itemize}
    \item We propose $\mathtt{FedBone}$, a novel federated multi-task learning framework via split learning for large-scale federated training on edge clients and heterogeneous task adaptation.
    \item We propose GPAggregation to alleviate optimization challenges of the general model posed by task heterogeneity among clients, which rescales client gradients with historical gradients attention and merge gradient conflict between clients.
    \item We conduct extensive experiments on two public multi-task datasets. The results show that our proposed $\mathtt{FedBone}$ outperforms existing federated learning algorithms in heterogeneous tasks with much smaller computational resource requirements. The experiments on 13 real-world Ophthalmic tasks reveal the potential capability of $\mathtt{FedBone}$ in real medical and healthcare applications.
\end{itemize}

\section{Related Work}

\subsection{Federated Multi-task Learning}
\begin{figure*}[htb]
    \centering
    \includegraphics[width=16cm]{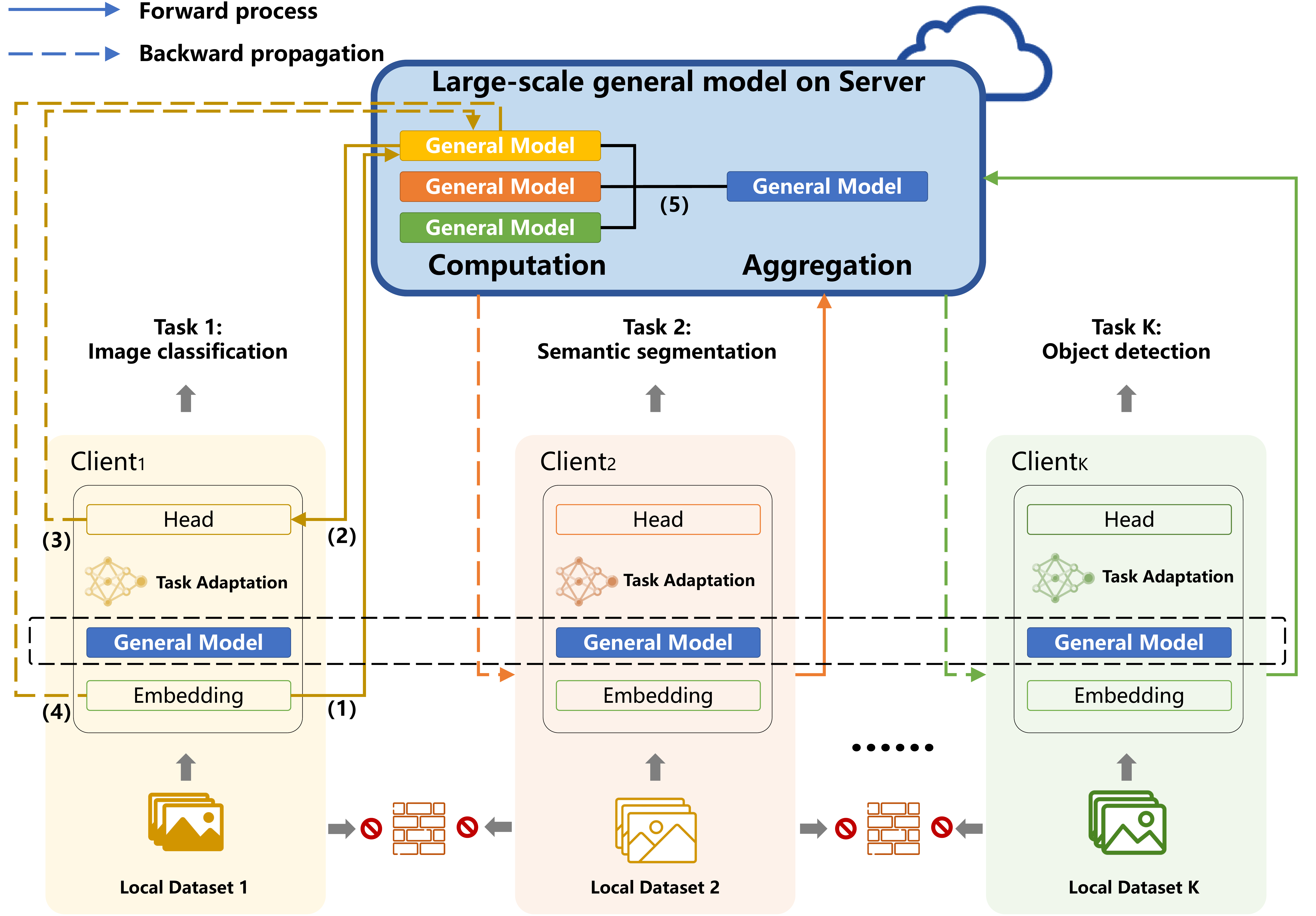}
    \caption{The workflow of $\mathtt{FedBone}$ framework. Clients perform patch embedding locally and (1) send embeddings to the cloud server for feature extraction using the general model, and the cloud server (2) sends extracted features back to clients. Clients complete the loss computation and (3) send backward intermediate results to the cloud server for backward propagation of the general model, and the cloud server (4) sends results back to clients. The clients can now update the local task adaptation module. For each client, the cloud server maintains a distinct general model, which is updated during every client's mini-batch. When all clients finish a local training epoch, the cloud server will (5) aggregate these general models to finalize one communication round.}
    \label{fig:overview}
\end{figure*}

Multi-task learning methods can be divided into centralized and distributed computing methods according to the data collection method. The former collects the data in advance to a central node and then runs the model~\cite{misra2016cross,kendall2018multi}, while the distributed method collects heterogeneous data from different tasks in a distributed way, but often faces the problems of high communication cost and privacy issues that prevent copying to the central node. Federated multi-task learning proposes to train different models directly on each node using knowledge sharing~\cite{smith2017federated}, focusing on the fault tolerance and privacy of the node datasets. The initial federated multi-task learning had a central node~\cite{smith2017federated}, and the latest research also achieved decentralized federated multi-task learning~\cite{marfoq2021federated}. Regularization-based multi-task learning methods capture complex relationships between personalized models to achieve aggregation of different tasks but lose the ability to grasp complex relationships between tasks~\cite{zantedeschi2020fully}. Ditto and other federated multi-task methods, although sacrificing the complexity of the regularization term to train more complex models, also lose the ability to capture complex relationships between tasks~\cite{li2021ditto}. The FATHOM framework leveraged the attention mechanism to extract input features and learn a shared temporal representation across different devices, thereby achieving knowledge transfer and performance improvement~\cite{chen2020federated}. FedMSplit was proposed to use a dynamic multi-view graph structure to address the modality incongruity problem among sensor devices and to promote local model relations through neighborhood message passing in the graph~\cite{chen2022fedmsplit}. The SpreadGNN framework has solved the non-I.I.D. problem of graph data and uses a dynamic multi-task optimization method to ensure model convergence~\cite{he2022spreadgnn}. FedICT achieved personalized services for multi-task clients by using federated prior distillation and local knowledge adjustment\cite{wu2023fedict}. However, none of the above approaches enables local training on clients with heterogeneous tasks, and they all require full model training and evaluating on local clients, which means that clients must equally have large computation capability when training large-scale models.

\subsection{Personalized federated learning}

In federated learning scenarios, a single global shared model faces the problem of significant differences in data distribution among clients. Personalized federated learning tries to solve this problem. There are two different strategies for personalized federated learning: personalization of the global model and individual personalized models. The former performs local adaptation for each client based on the trained global federated model to achieve personalized processing, while the latter trains individual personalized models on each client. In the data augmentation aspect, the self-balancing learning framework Astraea uses Z-score-based data augmentation and mediator-based multi-client rescheduling to mitigate the impact of data distribution differences~\cite{duan2020self}. In FedHome, each client performs personalized adaptation on a locally enhanced class-balanced dataset~\cite{wu2020fedhome}. Some studies use the method of adding a local loss regularization term, such as FedProx introduced an approximation term for the local subproblem, taking into account the dissimilarity between the global FL model and the local model to adjust the impact of local updates~\cite{li2020federated}, FedCL used the regularization term of elastic weight consolidation (EWC) from the continual learning domain\cite{yao2020continual}. Other methods such as transfer learning FedMd~\cite{li2019fedmd}, meta-learning MAML~\cite{jiang2019improving}, and so on are used to improve the performance of the global shared model trained on heterogeneous data in federated learning. The personalized solutions for clients mainly include methods such as parameter decoupling LG-FedAvg~\cite{liang2020think}, model interpolation HeteroFL~\cite{diao2020heterofl}, and clustering pFedBayes~\cite{zhang2022personalized}. Specifically, the importance of parameters in FedCurv was estimated by the Fisher information matrix, and a penalty step is performed to retain important parameters, which can reduce the catastrophic forgetting problem between multiple tasks~\cite{shoham2019overcoming}. The concept of personalized federated learning highlights the necessity of adapting a model for local data distribution. However, existing methods have failed to consider the potential for personalization in the event of heterogeneity of client tasks.

\section{Method}
\subsection{Problem Formulation}
We consider a set of federated clients $\mathcal{K}=\{1,2,...,K\}$, each client $k \in \mathcal{K}$ has a local dataset $\mathcal{D}_k=\{(x_{k}, y_{k})_i, i=1,2,...,N_k\}$ and collaboratively trains models with other federated learning clients, with the goal of training personalized local models $f_k$ that can adapt to the distinct local task. The goal is to solve the following optimization problems:
\begin{align}
    \forall k \in \mathcal{K},~~\min_{f_k \in \mathcal{F}} \mathcal{L}_k (f_k)
\end{align}
where $\mathcal{F}$ denotes the set of all personalized local models, $\mathcal{L}_k$ denotes local loss function.
%%本文关注于异构联邦多任务学习，参与方希望学习一个适配本地任务的模型，我们将本地模型笼统地分为数据嵌入$e$，特征提取$f$和任务输出$o$三部分。在预训练领域，我们最为关注的是中间的特征提取部分，这一部分也是计算量最高、能够提供的泛化性最强的部分，在假设参与方k本地
\subsection{Overall Architecture}
%%%%%%%%% 总体讲 %%%%%%%%%%%%%%%%
%% 人工智能模型能够适配多种多样的任务，这些任务可以归类为完整实体级别的分类、回归，也可能包括更细粒度和更复杂的任务，包括计算机视觉领域的语义分割、目标检测以及自然语言处理领域的序列标注与各种各样的生成任务等等。我们考虑在联邦学习中支持异构任务的参与方以纳入尽可能多的参与方进行大规模模型的联邦训练。我们提出的方法FedBone整体采用服务端计算大规模通用模型，边缘客户端计算数据嵌入和任务输出头的联邦分割学习方式。与直接平均的联邦聚合方式不同，FedBone采用了一种客户端梯度映射的方式聚合大规模通用模型，减小每轮联邦聚合过程中客户端之间的梯度冲突，以实现对大规模通用模型泛化性能的提升。考虑到异构任务的差异性，不同的任务关注的数据位置、数据尺度是不同的，可以认为大规模通用模型输出的隐层表示是全面但粗糙的，并不能直接应用于一些高级的深度学习任务。这里我们引入了结合可变形卷积和注意力机制组成的任务适配模块，通过可变形卷积适应形状不规则的特征，并通过自注意力机制捕捉通用特征表示中与任务相关的部分，最终提高参与方本地任务的性能。整体如图1所示。
% Deep learning models are capable of adapting to different tasks. These tasks can be low-level classification and regression tasks, as well as more high-level CV and NLP tasks, such as semantic segmentation and object detection in CV, and various generative tasks in NLP.
Our proposed framework $\mathtt{FedBone}$ aims to enable the participation of heterogeneous task clients in federated learning, thereby facilitating federated training of large-scale models. To achieve this, we adopt a split federated learning approach~\cite{thapa_splitfed_2022}, which involves the computation of a large-scale general model on the cloud server and lightweight computation of data embedding/task head output on the edge clients. $\mathtt{FedBone}$ aggregates large-scale general models using a task gradients projection method, which prevents gradient conflicts and improves model generalization performance, as opposed to the direct federated averaging aggregation methods. In order to enhance the performance of client local tasks, we introduce a task adaptation module, which comprises the deformable convolution and self-attention mechanism, that adapts to irregularly shaped feature maps through deformable convolution and captures the task interaction features. The full framework is illustrated in Figure \ref{fig:overview}.
In the following, we will outline the workflow of split federated multi-task learning and elaborate on the comprehensive design of federated aggregation via task gradient projection and task adaptation module.

\subsection{Split Federated Multi-Task Learning}
%%%%%%%%%%%%%%%%%%%%%%%%%%%%%%%
%%%%%%%%%%%% 讲Fed-Split %%%%%%%%%%%%%%
$\mathtt{FedBone}$ follows the split learning~\cite{thapa_splitfed_2022} approach, but it only requires one cloud server for high-performance computation and model aggregation. 
All clients perform patch embedding~\cite{liu_swin_2022} computations in parallel and then send the local results to the cloud server for feature extraction using a large-scale general model. After receiving client results, the cloud server responds to the client with general latent representations. Using these representations, clients then complete the task adaptation module and task output head forward propagation and immediately begin backward propagation. After the cloud server receives the gradients of the general model, it stores them and then sends the subsequent gradients of the task adaptation module to the original client. Clients with complete gradients can now update the parameters of the local patch embedding, task adaptation module, and task output head. When all selected clients send the gradients of the general model, the cloud server aggregates the gradients and updates the parameters of the general model. The specific detail can be found in Algorithm~\ref{alg:fedbone}.

In Algorithm \ref{alg:fedbone}, clients compute patch embedding $e(\cdot)$, task adaptation $l(\cdot)$ and task output head $o(\cdot)$ with paramenters $\zeta,\eta$ and $\phi$. The patch embedding module transposes raw data patches to flatten patch embeddings with a single convolution operation. The task adaptation module is built with deformable convolution and multi-head self-attention, which will be described in more detail in Section \ref{demt}. The task output head can vary for heterogeneous tasks, but it typically contains convolution, normalization, and deconvolution operations. Computation on clients yields relatively low resource requirements and can be conducted on low-power consumption edge devices. During clients update, the cloud server gradually gathers task gradients $\nabla^t_k$ for gradients aggregation and general model update subsequently.

\begin{algorithm}[tb]
    \caption{FedBone}
    \label{alg:fedbone}
    \textbf{Input}: Client set $\mathcal{K}$ with local datasets $\mathcal{D}_k, \forall k \in \mathcal{K}$ \\
    \textbf{Output}: General model $\theta$, client task-specific modules $\zeta,\eta,\phi$
    \begin{algorithmic}[1] %[1] enables line numbers
        \State Server initializes $\theta^0$, $\forall$ Client $k \in \mathcal{K}$ initializes $\zeta^0_k,\eta^0_k,\phi^0_k$
        \For{round $t=0,...,T-1$}
        \For{client $k=1,...,K$}
        \State Client patch embedding $x^t_{k,e}\gets e(x_k;\zeta^t_k)$
        \State Server feature extraction $x^t_{k,h}\gets f^(x^t_{k,e};\theta^t)$
        \State $\frac{\partial \mathcal{L}_k}{\partial f} \gets \Call{ClientUpdate}{x^t_{k,h}}$
        \State $\nabla^t_{k} = \frac{\partial \mathcal{L}_k}{\partial f} \frac{\partial f}{\theta^t}$
        \State Server send $\frac{\partial \mathcal{L}_k}{\partial e}$ to client
        \State Client completes backward propagation
        \State Client optimizes $\zeta^{t+1}_k,\eta^{t+1}_k,\phi^{t+1}_k$
        \EndFor
        \State Server gather $\nabla^t_{\mathcal{K}}=\{\nabla^t_{1}, \nabla^t_{2},..., \nabla^t_{K}\}$
        \State $\nabla^t \gets \Call{GPAggregation}{\nabla^t_{\mathcal{K}}, \nabla^{t-1}} $
        \State $\theta^{t+1} \gets \Call{Optimizer}{\theta^t,
        \nabla^t}$
        \EndFor

        \Function{ClientUpdate}{$x^t_{k,h}$}
        \State Task adaptation $x^t_{k,l} \gets l(x^t_{k,h};\eta^t_k)$
        \State Task output $\hat{y}^t_k \gets o(x^t_{k,l};\phi_t^k)$
        \State Task specific loss computation $\mathcal{L}_k(\hat{y}^t_k, y_k)$
        \State $\frac{\partial \mathcal{L}_k}{\partial f} \gets $ Backward propagation to task adaptation  
        \State \Return $\frac{\partial \mathcal{L}_k}{\partial f}$
        \EndFunction
    \end{algorithmic}
\end{algorithm}

\subsection{Gradients Aggregation via Conflicting Gradients Projection} \label{pcgrad}
\begin{figure}[htb]
    \centering
    \includegraphics[width=8cm]{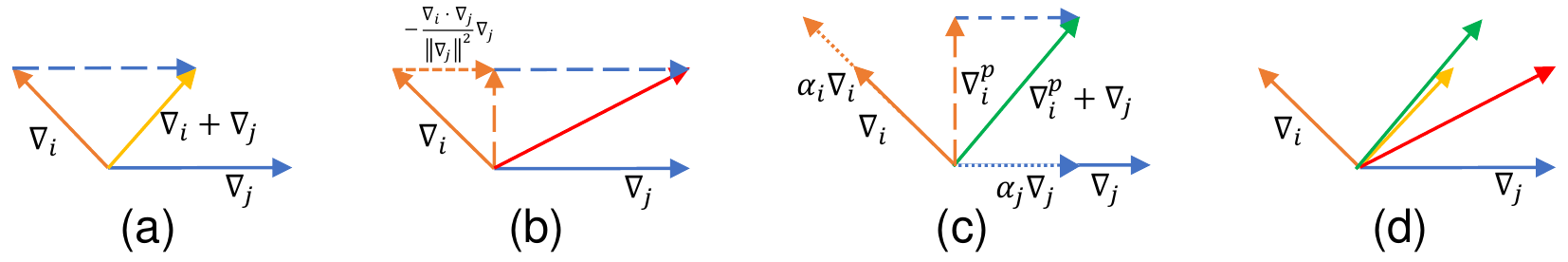}
    \caption{The gradients projection process of two task gradients $\nabla_i$ and $\nabla_j$. (a) The two gradients with conflicting gradient directions are aggregated directly, which can lead to interference. (b) The gradients $\nabla_i$ are firstly projected onto the normal vector of the gradients $\nabla_j$, and then they are aggregated. (c) The two gradients $\nabla_i$ and $\nabla_j$ are scaled with attention $\alpha$, and continue the projection-aggregation procedure. (d) The red, yellow, and green lines represent the aggregated gradients of the three cases, respectively.}
    \label{fig:pcgrad}
\end{figure}

%%%%%%%%%%%%% 讲conflicting理论，与联邦结合 %%%%%%%%%%%%%%%%%%%
%general model的更新依赖于云服务器上的梯度聚合。通过每个客户端计算到的loss反传可以得到general model梯度$\nabla^t_k$， 将所有客户端的梯度聚合可以优化general model的参数，提高genral model的generalization capability。
% 梯度与参数的聚合是联邦学习的一个重要研究领域【advances in fl引用】，在个性化【】、鲁棒性【】、公平性【】、计算效率【】、计算资源【】等多个方向上得到了广泛的研究并取得了大量进展。
% 在FedBone的框架下，参与方的本地任务要求是松弛的，任务具有异构性，优化目标和使用的损失函数也各异。在这种设定下直接平均聚合有可能降低general model的性能，【接已有的两句话】
The cloud server conducts gradient aggregation for optimizing parameters of the general model, which could integrate the knowledge of all client tasks and improve the generalization capability of the general model. Learning multiple tasks simultaneously is a challenging optimization problem that can sometimes lead to poorer model performance~\cite{rusu_policy_2016}. the optimization challenges are not well understood. In the federated learning scenario, things become even trickier, since most existing methods require access to raw data to build the relationship between tasks and determine the strategy for aggregating task gradients. To ease the need for raw data, one feasible approach is to attribute multi-objective optimization problems to the existence of gradient conflicts and described them as gradients from different tasks conflicting with one another among tasks \cite{yu2020gradient} and solve them by correcting the gradients. We now define conflicting gradients formally.
\begin{definition}[Conflicting gradients]
    Define the angle between client $i$ graidents $\nabla_i$ and client $j$ gradients $\nabla_j$ as $\omega_{ij}$, $\nabla_i$ and $\nabla_j$ are \textbf{conflicting gradients} when $\cos\omega_{ij}<0$.
\end{definition}
As shown in Figure \ref{fig:pcgrad}(a), gradients $\nabla_i$ and $\nabla_j$ have a negative impact on each other, and direct aggregation will cause a reduction in final gradients. An intuitive idea can be projecting one gradient $\nabla_i$ onto the normal plane of another gradient $\nabla_j$ with
\begin{align}
\nabla_i' = \nabla_i - \frac{\nabla_i \cdot \nabla_j}{\lVert \nabla_j \rVert^2}\nabla_{j}
\end{align}
to eliminate the opposite component, as shown in Figure \ref{fig:pcgrad}(b). The method works well when the general model converges towards flatter minima, but certain clients may fall into a sharp valley and the weight of the clients should be decreased in the aggregation procedure. To dampen the influence of clients which converge  towards sharp minima, we propose a novel gradients aggregation method \textbf{GPAggregation} by the use of historical aggregated gradients. We rescale gradients by calculating the attention values of historical aggregated gradients. A simple example is shown in Figure \ref{fig:pcgrad}(c), the gradient $\nabla_i$ is scaled by attention $\alpha_i$ and then projected onto the normal plane of scaled gradient $\nabla_j$.

%%%%%%%%%%%%%%%%%%%%%%%%%%%%%%%%%%%%%%%%%%%%%%%%%%

%%%%%%%%%%%%%% 讲PCGrad算法 %%%%%%%%%%%%%%%%%%%%%
The gradient projection method is described in Algorithm \ref{alg:pcgrad}. The task gradients $\nabla_k$ are scaled by attention mechanism with historical aggregated gradients $\nabla'$:
\begin{align}
    \nabla_k=softmax(\frac{\nabla_{k}\nabla'^T}{d_{\nabla}})\nabla_{k}.
\end{align}

Iterate through gradients of all other clients and project onto every normal plate, we now get the de-conflicted task gradients which can be used for average aggregation.
%%%%%%%%%%%%%%%%%%%%%%%%%%%%%%%%%%%%%%%%%%%%%%%
\begin{algorithm}[tb]
    \caption{GPAggregation}
    \label{alg:pcgrad}
    \textbf{Input}: A set of current round task gradients $\nabla_{\mathcal{K}}$, previous round aggregated gradients $\nabla'$\\
    \textbf{Output}: Aggregated gradients $\nabla$
    \begin{algorithmic}[1] %[1] enables line numbers
        \For{$\nabla_{k} \in \nabla_{\mathcal{K}}$}
            \State $\nabla_{k} = \Call{Softmax}{\frac{\nabla_{k}\nabla'^T}{d_{\nabla}}}\nabla_{k}$
        \EndFor
        \State store $\nabla^p_k \gets \nabla_k, \forall \nabla_k ~in~ \nabla_{\mathcal{K}}$
        \For{$\nabla_{k} \in \nabla_{\mathcal{K}}$}
            \For{$\nabla_{i} \in \nabla_{\mathcal{K}} \setminus \nabla_{k}$}
                \If{$\nabla^p_{k} \cdot \nabla_{i} < 0$}
                    \State $\nabla^p_{k} = \nabla^p_{k} - \frac{\nabla^p_{k} \cdot \nabla_{i}}{\lVert \nabla_{i} \rVert^2}\nabla_{i}$
                \EndIf
            \EndFor
        \EndFor
        \State $\nabla = \frac{1}{K}\sum_k{\nabla^p_{k}}$
    \end{algorithmic}
\end{algorithm}

\subsection{Heterogeneous Task Adaptation} \label{demt}
%%%%%%%%%%%%% motivation %%%%%%%%%%%%%%%%%%%

%%%%%%%%%%%%%%%%%%%%%%%%%%%%%%%%%%%%%%%%%%%%%%%%%%
%%%%%%%%%%%%%% Defor+Attention模块结构 %%%%%%%%%%%%%%%%%%%%%
The large-scale general model can extract latent representations with sufficient information for handling heterogeneous tasks on the client side. In general multi-task learning, similar task output header structures are used for different tasks to reduce the complexity of optimizing model parameters~\cite{zamir_taskonomy_2018}. In a federated multi-task learning scenario, distribution shifts exist among clients. more unevenly than centralized multi-task learning data. As a result, the latent representations by the general model are more generalized and decoupled from the specific distribution of client data. Relying solely on a lightweight task output head makes it challenging to extract further task-specific information from the general latent representations and apply it to accomplish tasks. leading to a more obvious distribution shift. Inspired by the successful Deformable Convolutional Network~\cite{dai_deformable_2017} and Convolutional Transformer joint structure~\cite{xu2023demt}, we propose a heterogeneous task adaptation module that adaptively captures unique receptive regions specific to each task and task interactions. The heterogeneous task adaptation module uses channel-wise pooling, spatial-wise sampling, and intra-task attention to learn relevant task-specific features. Utilizing the reconstructed feature representations enables the task output head to perform downstream tasks more effectively and efficiently.

\begin{figure}[htb]
    \centering
    \includegraphics[width=8cm]{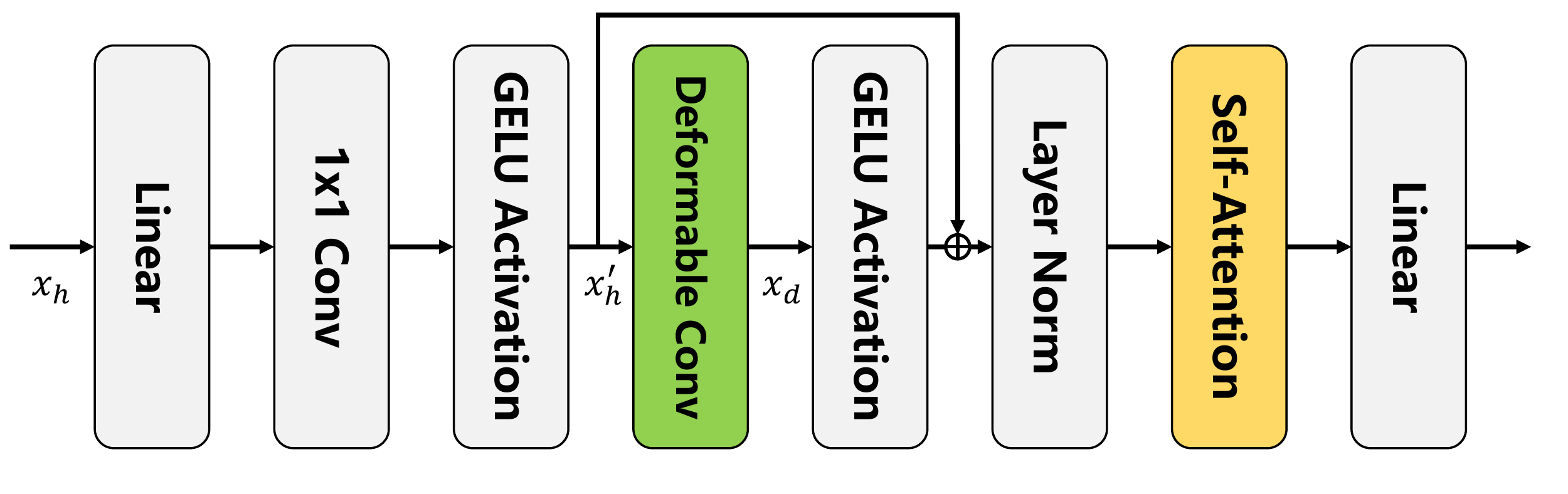}
    \caption{Illustration of task adaptation module}
    \label{fig:demt}
\end{figure}

%%%%%%%%%%%%%%% 对模块具体介绍 %%%%%%%%%%%%%%%%%%%%%%%%
% 5.14 to huang "To generate the relative offsets"之前还需要介绍一下可变形卷积的基本信息，然后引出要算offfset这块，最后再加上residual连接和multi-head self-attention计算输出就行了 可以放一个自注意力的公式
% ^^^^^^^^^^^^^^^^^^^^^^^^^^^^^^^^^......
As shown in Figure \ref{fig:demt}, the heterogeneous task adaptation module mainly consists of $1\times1$ convolution, deformable convolution and self-attention mechanism. The module uses general latent representation $x_h$ received from the cloud server which is initially fed into a linear layer to reduce the channel dimension. The feature map then employs $1\times1$ convolution to communicate between channels. Following the GELU activation, the resulting feature map is denoted as $x_h'$. 

Following~\cite{dai_deformable_2017}, we first sample a regular grid $R$ over the input feature map $x_h'$ and then the summation of sampled values weighted by $w$. To generate the relative offsets with respect to the reference point $\mathbf{p}$, the full feature map $x_h'$ is fed to the convolution operator to learn the corresponding offsets $\delta_{\mathbf{p}}$. %切入可变形卷积% For each location point $\mathbf{p}$ and a regular grid $\mathcal{R}$, the deformable convolution can be formulated as:
\begin{align}
    x_d(\mathbf{p}) = \sum_{\delta_{\mathbf{p}} \in \mathcal{R}} w(\mathbf{p}) \cdot x_h'(\mathbf{p}+\delta_{\mathbf{p}}).
\end{align}

% The output is then projected $x_d$  into the queries ($Q$), keys ($K$) and values ($V$) of dimension $d_k$ using linear transformation. The self-attention block is :
% \begin{align}
%     x = {\rm softmax}\left(\dfrac{QK^T}{\sqrt{d_k}}\right)V,
% \end{align}

%%%%%%%%%%%%%%%%%%%%%%%%%%%%%%%%%%%%%%%%%%%%%%%

\begin{table*}[htb]
\centering
\tiny
\setlength{\tabcolsep}{4pt}
\begin{tabular}{@{}lllllllllllllllllllll@{}}
\toprule
\multirow{2}{*}{Method} & \multicolumn{5}{c}{Segmentation(mIOU)$\uparrow$}                                                                         & \multicolumn{5}{c}{Depth(RMSE)$\downarrow$}                                                                             & \multicolumn{5}{c}{Normal(mErr)$\downarrow$}                                                                            & \multicolumn{5}{c}{Bound(osdF)$\uparrow$}                                                                               \\
                        & \multicolumn{1}{c}{1} & \multicolumn{1}{c}{2} & \multicolumn{1}{c}{3} & \multicolumn{1}{c}{4} & \multicolumn{1}{c}{Avg} & \multicolumn{1}{c}{1} & \multicolumn{1}{c}{2} & \multicolumn{1}{c}{3} & \multicolumn{1}{c}{4} & \multicolumn{1}{c}{Avg} & \multicolumn{1}{c}{1} & \multicolumn{1}{c}{2} & \multicolumn{1}{c}{3} & \multicolumn{1}{c}{4} & \multicolumn{1}{c}{Avg} & \multicolumn{1}{c}{1} & \multicolumn{1}{c}{2} & \multicolumn{1}{c}{3} & \multicolumn{1}{c}{4} & \multicolumn{1}{c}{Avg} \\ \midrule
FedAvg                  & 38.97                 & 38.29                 & \textbf{43.30}        & 37.48                 & 39.51                   & 0.4283                & 0.5294                & 0.5571                & 0.7170                & 0.5580                  & 27.01                 & 26.52                 & 25.64                 & 28.40                 & 26.89                   & 62.57                 & 62.23                 & 63.67                 & 59.86                 & 62.08                   \\
FedProx                 & 29.25                 & 31.92                 & 31.68                 & 24.66                 & 29.38                   & 0.4846                & 0.6069                & 0.6155                & 0.7892                & 0.6241                  & 27.68                 & 27.33                 & 26.12                 & 28.75                 & 27.47                   & 61.01                 & 61.27                 & 62.70                 & 58.15                 & 60.78                   \\
pFedMe                  & 33.14                 & 33.32                 & 33.90                 & 26.74                 & 31.78                   & 0.4340                & 0.5314                & 0.5466                & 0.7457                & 0.5644                  & 23.93                 & 25.97                 & 22.76                 & \textbf{27.50}        & 25.04                   & 62.29                 & 62.92                 & 64.57                 & 60.77                 & 62.63                   \\
FedEM                   & 41.57                 & 38.33                 & 41.97                 & 40.31                 & 40.55                   & \textbf{0.4023}                & 0.5221                & \textbf{0.5346}       & 0.7215                & 0.5451                  & 26.44                 & 26.41                 & 25.11                 & 28.16                 & 26.53                   & 62.81                 & \textbf{63.91}        & 63.82                 & 60.43                 & 62.74                   \\ \midrule
\textbf{FedBone(ours)}  & \textbf{42.92}        & \textbf{40.73}        & 43.22                 & \textbf{42.47}        & \textbf{42.34}          & 0.4594                & \textbf{0.5190}       & 0.5407                & \textbf{0.6136}       & \textbf{0.5332}         & \textbf{23.67}        & \textbf{25.32}        & \textbf{22.57}        & 27.78                 & \textbf{24.84}          & \textbf{63.46}        & 63.74                 & \textbf{65.04}        & \textbf{61.32}        & \textbf{63.39}          \\ \bottomrule
\end{tabular}
\caption{Comparison of FL methods on NYUDv2 dataset, $\uparrow$ means higher is better, and $\downarrow$ means lower is better.}
\label{tab:nyud}
\end{table*}

\section{Experiments}
\subsection{Experimental Setup}
\subsubsection{Implementation}
We evaluate the performance of $\mathtt{FedBone}$ on two multi-task dense prediction datasets and compare the results with common FL method FedAvg~\cite{mcmahan_communication-efficient_2017}, personalized FL methods FedProx~\cite{li_federated_2020} and pFedMe~\cite{dinh_personalized_2020}, and multi-task FL method FedEM~\cite{marfoq2021federated}.
% 蒋客户端划分，通信，和模型结构
We determine the number of clients by the tasks of each dataset. For each task in the dataset, we randomly split it into 4 clients with equal data volumes, and partitioned the training and testing sets in a 8:2 ratio on the clients. For the FL methods used for comparison, we designed a fully convolutional~\cite{long_fully_2015} task-specific output head for each task. For every FL method, we set communication rounds to 200.
% 超参数，优化器，其他
For $\mathtt{FedBone}$, FedAvg and FedEM, we use SGD as the optimizer. For FedProx and pFedMe, we have modified the SGD optimizer to fit the optimization process of the algorithm. The batch size is set to 16 and the learning rate is set to 0.01, scheduled to decay by a fraction of 0.1 every 50 epochs. All our experiments are conducted on the Pytorch framework with 8 NVIDIA A800 80GB GPUs and 1TB system memory.

\subsubsection{Datasets}
We adopt two publicly accessible datasets NYUDv2~\cite{silberman_indoor_2012} and PASCAL-Context~\cite{mottaghi_role_2014}. NYUDv2 contains 1,449 RGB images and provides dense labels for semantic segmentation, depth estimation, normal estimation, and boundary detection tasks. PASCAL-Context contains 10,180 training RGB images with dense labels for semantic segmentation, saliency estimation, normal estimation, and boundary detection tasks. Meanwhile, Xu~\shortcite{chen_detect_2014} provides extra human parts annotations for 3,589 images, which act as the labels for the human part segmentation task.

\subsubsection{Metrics}
The chosen datasets comprise a total of 5 different types of tasks. For semantic segmentation tasks (including human part segmentation), we use mean Intersection over Union (mIoU) as the metric. For normal estimation tasks, mean Error (mErr) is adopted, and for boundary detection tasks, optimal dataset scale F-measure (odsF) is used. For 
 the depth and saliency estimation, the Root Mean Square Error (RMSE) and the maximum F-measure (maxF) are exploited respectively.

\subsubsection{Backbones}
We employ Swin Transformer Small (Swin-S) ~\cite{liu_swin_2022} pre-trained on ImageNet-22K~\cite{mmseg2020} as the backbone for all experiments except the analysis of computational resource requirements. In order to accommodate the demand of model scale in the production environment, we used a larger model Swin Transformer Base (Swin-B)  as the backbone to analyze the computational and memory resources required by various FL methods on the client side.
\begin{table}[htb]
    \centering
    \small
    \setlength{\tabcolsep}{2pt}
    \begin{tabular}{@{}lccccc@{}}
    \toprule
    Method                 & \makecell{Segment \\ (mIOU)$\uparrow$} & \makecell{HumanPart \\ (mIOU)$\uparrow$} & \makecell{Saliency \\ (maxF)$\uparrow$} & \makecell{Normal \\ (mErr)$\downarrow$} & \makecell{Bound \\ (odsF)$\uparrow$} \\ \midrule
    FedAvg                 & 52.71                  & 56.12                    & 82.97                    & 17.66                    & 61.23                 \\
    FedProx                & 61.69                  & 53.21                    & 81.48                    & 15.69                    & 62.32                 \\
    pFedMe                 & 59.16                  & 57.04                    & 80.90                    & 15.67                    & 66.59                 \\
    FedEM                  & 51.10                  & 53.79                    & 82.15                    & 19.64                    & 59.27                 \\ \midrule
    \textbf{FedBone(ours)} & \textbf{62.74}         & \textbf{58.09}           & \textbf{84.36}           & \textbf{15.13}           & \textbf{66.42}        \\ \bottomrule
    \end{tabular}
    \caption{Comparison of FL methods on PASCAL-Context dataset}
    \label{tab:pascal}
\end{table}
\subsection{Dense Prediction Tasks}
Table 1 and Table 2 present the performance of $\mathtt{FedBone}$ on the NYUDv2 dataset and the PASCAL-Context dataset. These tables compare the performance of four different methods, namely FedAvg, FedProx, pFedMe, and FedEM, across multiple tasks. In certain tasks such as segmentation, humanpart, saliency, and bound, higher values indicate better performance, whereas in tasks like depth and normal, lower values indicate superior performance. Our method, $\mathtt{FedBone}$, consistently outperforms all the comparative methods in the segmentation, humanpart, saliency, and bound tasks, demonstrating its superiority with higher metric values. Conversely, for the depth and normal tasks, $\mathtt{FedBone}$ consistently achieves lower values, indicating its better performance compared to the comparative methods. These results highlight the effectiveness and generalization of $\mathtt{FedBone}$ across diverse tasks. 

Overall, $\mathtt{FedBone}$ exhibits clear advantages over the comparative methods in terms of average accuracy. While there are a few isolated cases where $\mathtt{FedBone}$, trained on a single client, falls slightly short compared to the comparative methods, it consistently outperforms them when considering the overall average accuracy. This underscores the significant advantage of our approach in terms of overall performance. The findings demonstrate the effectiveness of $\mathtt{FedBone}$ in achieving superior performance across multiple tasks, validating its potential as a robust method in federated learning scenarios.

\begin{table}[htb]
    \setlength{\tabcolsep}{4pt}
    \centering
    \small
    \begin{tabular}{@{}lccc@{}}
    \toprule
    Methods       & \makecell{Parameters \\ (m)}       & \makecell{GFLOPS \\ (G)}            & \makecell{Memory \\ (GB)} \\ \midrule
    FedAvg        & 95.04               & 123.30               & 33.32   \\
    FedProx       & 95.04               & 123.30               & 33.68   \\
    pFedMe        & 95.04               & 123.30               & 33.68   \\
    FedEM         & 285.13              & 369.91               & 36.13   \\
    \textbf{FedBone(ours)} & \textbf{1.92}(+88.67) & \textbf{11.74}(+87.05) & \textbf{3.31}(+32.12)    \\ \bottomrule
    \end{tabular}
    \caption{Computational resources required by FL methods on the client-side when training the Swin-B model. The + number in brackets means the required resources on the cloud server.}
    \label{tab:resource}
\end{table}
\begin{table*}[htb]
    \centering
    \setlength{\tabcolsep}{4pt}
    \begin{tabular}{@{}lccccccccccccc@{}}
    \toprule
    Methods       & HMM            & RVO            & RP             & DME            & PM             & FE             & HR             & G              & MEM            & MH             & AMD            & DR             & LS(mIOU)        \\ \midrule
    FedAvg        & 94.33          & 93.72          & 96.07          & 93.52          & 91.27          & 78.27          & 94.35          & 94.22          & 92.25          & 94.56          & 79.63          & 63.86          & 49.71          \\
    FedProx       & 97.91          & 97.86          & 94.3           & 94.61          & 92.36          & 80.2           & 93.20          & 95.88          & 94.58          & 97.64          & 88.47          & 92.98          & 50.69          \\
    pFedMe        & 98.62          & 96.57          & 94.94          & 98.01          & 92.88          & 81.3           & 94.17          & 96.08          & 95.99          & 97.68          & 90.31          & 92.06          & 52.63          \\
    FedEM         & 98.79          & \textbf{99.41} & 96.14          & 96.22          & 91.25          & 83.3           & 95.23          & 96.22          & 95.92          & 97.03          & 90.15          & \textbf{94.68} & 54.21          \\ \midrule
    \textbf{FedBone(Ours)} & \textbf{98.87} & 98.91          & \textbf{99.24} & \textbf{99.13} & \textbf{93.71} & \textbf{84.57} & \textbf{95.90} & \textbf{96.75} & \textbf{96.15} & \textbf{98.93} & \textbf{91.18} & 94.57          & \textbf{55.82} \\ \bottomrule
    \end{tabular}
    \caption{Comparison of FL methods on real ophthalmic dataset}
    \label{tab:ophthalmic}
\end{table*}
\subsection{Analysis}
\subsubsection{Ablation Study}
\begin{figure}[t]
\includegraphics[width=\linewidth]{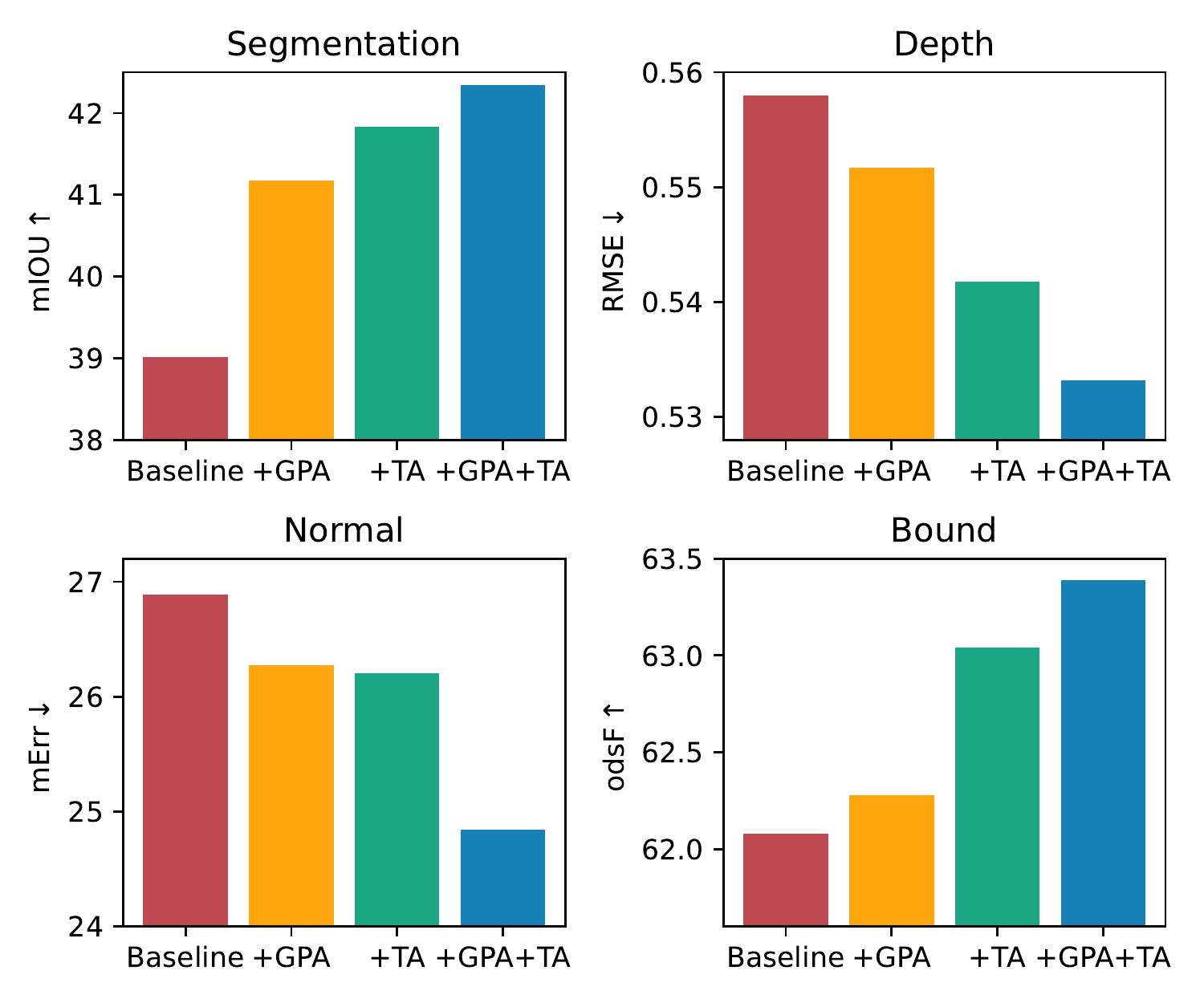}
\caption{The overview of our proposed framework}
\label{fig:ablation}
\end{figure}

We conducted an ablation study of $\mathtt{FedBone}$ to evaluate the contribution of each component and setting. The results are presented in Figure \ref{fig:ablation}. 

The baseline FedAvg is shown in the first row, while the +GPA and +TA indicate the addition of the GPAggregation and heterogeneous task adaptation module to the baseline. The table shows that compared to FedAvg, the addition of either the GPAggregation or task adaptation module results in improved performance, with the task adaptation module providing a more significant gain. This finding supports that heterogeneity between different tasks is a critical factor to consider when applying federated learning across tasks. The last bar of Figure \ref{fig:ablation} shows the performance of the proposed $\mathtt{FedBone}$. By integrating both the GPAggregation and task adaptation module, $\mathtt{FedBone}$ achieves the best performance among the different settings evaluated.

% \begin{table}[htb]
%     \setlength{\tabcolsep}{4pt}
%     \centering
%     \begin{tabular}{@{}cllll@{}}
%     \toprule
%     Modules          & \makecell{Segment \\ (mIoU)} & \makecell{Depth \\ (rmse)} & \makecell{Normal \\ (mErr)} & \makecell{Bound \\ (odsF)} \\ \midrule
%     Baseline(FedAvg) & 39.01    & 0.5580      & 26.89        & 62.08       \\
%     $+$ GPA           & 41.17    & 0.5517      & 26.27        & 62.28       \\
%     $+$ TA            & 41.83    & 0.5418      & 26.20        & 63.04       \\
%     $+$ GPA and TA(ours)  & 42.34    & 0.5332      & 24.84        & 63.39       \\ \bottomrule
%     \end{tabular}
%     \caption{Ablation on modules}
%     \label{tab:ablation}
%     \end{table}
\subsubsection{Computational Resource Requirements}
We conduct an analysis of the computational resource requirements to compare $\mathtt{FedBone}$ with other FL methods. In Table \ref{tab:resource}, FL methods FedAvg, FedProx, and pFedMe, which employ the fully convolutional task-specific head, have a total parameter number similar to that of $\mathtt{FedBone}$. However, $\mathtt{FedBone}$ utilizes the split learning paradigm, which places most computations on the cloud server, and thus, the majority of parameters are not stored locally, resulting in a vast disparity in local computation and memory usage during training. The FedEM implements ensemble learning and has triple the parameters of common FL methods. Nevertheless, the total memory usage is comparable since it trains sequentially in effect.

\subsection{Real-world Ophthalmic Tasks}
To further investigate the effectiveness of our proposed method $\mathtt{FedBone}$ in real-world applications, we collect 12,912 color fundus images and label the images according to ophthalmic diseases, including high myopia maculopathy (HMM), retinal vein occlusion (RVO), proliferative retinopathy (PR), diabetic macular edema (DME), pathological myopia (PM), hypertensive retinopathy (HR), glaucoma (G), macular epiretinal membrane (MEM), macular hole (MH).
We label images that show potential pathological changes but could not be diagnosed as any specific disease, as needing further examination (FE). The ten diseases, combined with the normal fundus labels, form 10 binary classification (disease diagnosis) tasks.
In addition to labeling for disease diagnosis, we also conduct labeling for two types of disease grading, i.e., age-related macular degeneration (AMD) grading and diabetic retinopathy (DR) grading. Together with Retinal-Lesions~\cite{wei_learn_2021} retinal lesion segmentation dataset, we build up a 13-task real-world ophthalmic dataset, the results are shown in Table \ref{tab:ophthalmic}.

All FL methods perform well on simple binary classification tasks in Table \ref{tab:ophthalmic}. Overall, personalized FL methods, including FedProx, pFedMe, and FedEM, perform better than the common FL method FedAvg. Additionally, our proposed $\mathtt{FedBone}$ achieved the best performance on the vast majority of tasks. For the ophthalmic semantic segmentation task LS, $\mathtt{FedBone}$ outperforms all other FL methods, which shows the potential of $\mathtt{FedBone}$ in real medical scenarios.

\section{Conclusion}
In this paper, we proposed a novel federated multi-task learning framework $\mathtt{FedBone}$  via split learning for large-scale federated training on edge clients and heterogeneous task adaptation.
To enhance the general model’s generalization, we introduce an aggregation method GPAggregation, which rescales client gradients with attention to historical gradients and merges gradient conflict between clients. The extensive experiments show that $\mathtt{FedBone}$ outperforms existing federated learning algorithms in heterogeneous tasks with off-the-shelf computational resources on the client side. The real ophthalmic experiment also indicates a promising future in using $\mathtt{FedBone}$ for real medical and healthcare applications. In the future, we may further extend $\mathtt{FedBone}$ for more data modality and reduce the communication cost.

\section*{Acknowledgments}
This work is supported by Beijing Municipal Science \& Technology Commission (No.Z221100002722009), National Natural Science Foundation of China (No.62202455), Youth Innovation Promotion Association CAS, and the Science Research Foundation of the Joint Laboratory Project on Digital Ophthalmology and Vision Science (No. SZYK202201).

\bibliographystyle{named}
\bibliography{fl_ijcai23}

\begin{thebibliography}{}

\bibitem[\protect\citeauthoryear{Cao \bgroup \em et al.\egroup
  }{2023}]{cao_cross-silo_2023}
Xingjian Cao, Zonghang Li, Gang Sun, Hongfang Yu, and Mohsen Guizani.
\newblock Cross-silo heterogeneous model federated multitask learning.
\newblock {\em Knowledge-Based Systems}, 265:110347, April 2023.

\bibitem[\protect\citeauthoryear{Chen and Zhang}{2022}]{chen2022fedmsplit}
Jiayi Chen and Aidong Zhang.
\newblock Fedmsplit: Correlation-adaptive federated multi-task learning across
  multimodal split networks.
\newblock In {\em Proceedings of the 28th ACM SIGKDD Conference on Knowledge
  Discovery and Data Mining}, pages 87--96, 2022.

\bibitem[\protect\citeauthoryear{Chen \bgroup \em et al.\egroup
  }{2014}]{chen_detect_2014}
Xianjie Chen, Roozbeh Mottaghi, Xiaobai Liu, Sanja Fidler, Raquel Urtasun, and
  Alan Yuille.
\newblock Detect {What} {You} {Can}: {Detecting} and {Representing} {Objects}
  using {Holistic} {Models} and {Body} {Parts}.
\newblock In {\em Proceedings of the {IEEE} {Conference} on {Computer} {Vision}
  and {Pattern} {Recognition}}, pages 1971--1978, 2014.

\bibitem[\protect\citeauthoryear{Chen \bgroup \em et al.\egroup
  }{2020}]{chen2020federated}
Yujing Chen, Yue Ning, Zheng Chai, and Huzefa Rangwala.
\newblock Federated multi-task learning with hierarchical attention for sensor
  data analytics.
\newblock In {\em 2020 International Joint Conference on Neural Networks
  (IJCNN)}, pages 1--8. IEEE, 2020.

\bibitem[\protect\citeauthoryear{Contributors}{2020}]{mmseg2020}
MMSegmentation Contributors.
\newblock {MMSegmentation}: Openmmlab semantic segmentation toolbox and
  benchmark.
\newblock \url{https://github.com/open-mmlab/mmsegmentation}, 2020.

\bibitem[\protect\citeauthoryear{Dai \bgroup \em et al.\egroup
  }{2017}]{dai_deformable_2017}
Jifeng Dai, Haozhi Qi, Yuwen Xiong, Yi~Li, Guodong Zhang, Han Hu, and Yichen
  Wei.
\newblock Deformable {Convolutional} {Networks}.
\newblock In {\em Proceedings of the {IEEE} {International} {Conference} on
  {Computer} {Vision}}, pages 764--773, 2017.

\bibitem[\protect\citeauthoryear{Diao \bgroup \em et al.\egroup
  }{2020}]{diao2020heterofl}
Enmao Diao, Jie Ding, and Vahid Tarokh.
\newblock Heterofl: Computation and communication efficient federated learning
  for heterogeneous clients.
\newblock {\em arXiv preprint arXiv:2010.01264}, 2020.

\bibitem[\protect\citeauthoryear{Dinh \bgroup \em et al.\egroup
  }{2020}]{dinh_personalized_2020}
Canh~T. Dinh, Nguyen~H. Tran, and Tuan~Dung Nguyen.
\newblock Personalized federated learning with moreau envelopes.
\newblock In {\em Proceedings of the 34th {International} {Conference} on
  {Neural} {Information} {Processing} {Systems}}, {NIPS}'20, pages
  21394--21405, Red Hook, NY, USA, December 2020. Curran Associates Inc.

\bibitem[\protect\citeauthoryear{Duan \bgroup \em et al.\egroup
  }{2020}]{duan2020self}
Moming Duan, Duo Liu, Xianzhang Chen, Renping Liu, Yujuan Tan, and Liang Liang.
\newblock Self-balancing federated learning with global imbalanced data in
  mobile systems.
\newblock {\em IEEE Transactions on Parallel and Distributed Systems},
  32(1):59--71, 2020.

\bibitem[\protect\citeauthoryear{He \bgroup \em et al.\egroup
  }{2022}]{he2022spreadgnn}
Chaoyang He, Emir Ceyani, Keshav Balasubramanian, Murali Annavaram, and Salman
  Avestimehr.
\newblock Spreadgnn: Decentralized multi-task federated learning for graph
  neural networks on molecular data.
\newblock In {\em Proceedings of the AAAI Conference on Artificial
  Intelligence}, volume~36, pages 6865--6873, 2022.

\bibitem[\protect\citeauthoryear{Jiang \bgroup \em et al.\egroup
  }{2019}]{jiang2019improving}
Yihan Jiang, Jakub Kone{\v{c}}n{\`y}, Keith Rush, and Sreeram Kannan.
\newblock Improving federated learning personalization via model agnostic meta
  learning.
\newblock {\em arXiv preprint arXiv:1909.12488}, 2019.

\bibitem[\protect\citeauthoryear{Kendall \bgroup \em et al.\egroup
  }{2018}]{kendall2018multi}
Alex Kendall, Yarin Gal, and Roberto Cipolla.
\newblock Multi-task learning using uncertainty to weigh losses for scene
  geometry and semantics.
\newblock In {\em Proceedings of the IEEE conference on computer vision and
  pattern recognition}, pages 7482--7491, 2018.

\bibitem[\protect\citeauthoryear{Li and Wang}{2019}]{li2019fedmd}
Daliang Li and Junpu Wang.
\newblock Fedmd: Heterogenous federated learning via model distillation.
\newblock {\em arXiv preprint arXiv:1910.03581}, 2019.

\bibitem[\protect\citeauthoryear{Li \bgroup \em et al.\egroup
  }{2020a}]{li2020federated}
Tian Li, Anit~Kumar Sahu, Manzil Zaheer, Maziar Sanjabi, Ameet Talwalkar, and
  Virginia Smith.
\newblock Federated optimization in heterogeneous networks.
\newblock {\em Proceedings of Machine learning and systems}, 2:429--450, 2020.

\bibitem[\protect\citeauthoryear{Li \bgroup \em et al.\egroup
  }{2020b}]{li_federated_2020}
Tian Li, Anit~Kumar Sahu, Manzil Zaheer, Maziar Sanjabi, Ameet Talwalkar, and
  Virginia Smith.
\newblock Federated {Optimization} in {Heterogeneous} {Networks}.
\newblock {\em Proceedings of Machine Learning and Systems}, 2:429--450, March
  2020.

\bibitem[\protect\citeauthoryear{Li \bgroup \em et al.\egroup
  }{2021}]{li2021ditto}
Tian Li, Shengyuan Hu, Ahmad Beirami, and Virginia Smith.
\newblock Ditto: Fair and robust federated learning through personalization.
\newblock In {\em International Conference on Machine Learning}, pages
  6357--6368. PMLR, 2021.

\bibitem[\protect\citeauthoryear{Liang \bgroup \em et al.\egroup
  }{2020}]{liang2020think}
Paul~Pu Liang, Terrance Liu, Liu Ziyin, Nicholas~B Allen, Randy~P Auerbach,
  David Brent, Ruslan Salakhutdinov, and Louis-Philippe Morency.
\newblock Think locally, act globally: Federated learning with local and global
  representations.
\newblock {\em arXiv preprint arXiv:2001.01523}, 2020.

\bibitem[\protect\citeauthoryear{Liu \bgroup \em et al.\egroup
  }{2022}]{liu_swin_2022}
Ze~Liu, Han Hu, Yutong Lin, Zhuliang Yao, Zhenda Xie, Yixuan Wei, Jia Ning, Yue
  Cao, Zheng Zhang, Li~Dong, Furu Wei, and Baining Guo.
\newblock Swin {Transformer} {V2}: {Scaling} {Up} {Capacity} and {Resolution}.
\newblock In {\em 2022 {IEEE}/{CVF} {Conference} on {Computer} {Vision} and
  {Pattern} {Recognition} ({CVPR})}, pages 11999--12009, June 2022.
\newblock ISSN: 2575-7075.

\bibitem[\protect\citeauthoryear{Long \bgroup \em et al.\egroup
  }{2015}]{long_fully_2015}
Jonathan Long, Evan Shelhamer, and Trevor Darrell.
\newblock Fully {Convolutional} {Networks} for {Semantic} {Segmentation}.
\newblock In {\em Proceedings of the {IEEE} {Conference} on {Computer} {Vision}
  and {Pattern} {Recognition}}, pages 3431--3440, 2015.

\bibitem[\protect\citeauthoryear{Marfoq \bgroup \em et al.\egroup
  }{2021}]{marfoq2021federated}
Othmane Marfoq, Giovanni Neglia, Aur{\'e}lien Bellet, Laetitia Kameni, and
  Richard Vidal.
\newblock Federated multi-task learning under a mixture of distributions.
\newblock {\em Advances in Neural Information Processing Systems},
  34:15434--15447, 2021.

\bibitem[\protect\citeauthoryear{McMahan \bgroup \em et al.\egroup
  }{2017}]{mcmahan_communication-efficient_2017}
Brendan McMahan, Eider Moore, Daniel Ramage, Seth Hampson, and Blaise Aguera~y
  Arcas.
\newblock Communication-{Efficient} {Learning} of {Deep} {Networks} from
  {Decentralized} {Data}.
\newblock In {\em Artificial {Intelligence} and {Statistics}}, pages
  1273--1282. PMLR, April 2017.
\newblock ISSN: 2640-3498.

\bibitem[\protect\citeauthoryear{Misra \bgroup \em et al.\egroup
  }{2016}]{misra2016cross}
Ishan Misra, Abhinav Shrivastava, Abhinav Gupta, and Martial Hebert.
\newblock Cross-stitch networks for multi-task learning.
\newblock In {\em Proceedings of the IEEE conference on computer vision and
  pattern recognition}, pages 3994--4003, 2016.

\bibitem[\protect\citeauthoryear{Mottaghi \bgroup \em et al.\egroup
  }{2014}]{mottaghi_role_2014}
Roozbeh Mottaghi, Xianjie Chen, Xiaobai Liu, Nam-Gyu Cho, Seong-Whan Lee, Sanja
  Fidler, Raquel Urtasun, and Alan Yuille.
\newblock The {Role} of {Context} for {Object} {Detection} and {Semantic}
  {Segmentation} in the {Wild}.
\newblock In {\em 2014 {IEEE} {Conference} on {Computer} {Vision} and {Pattern}
  {Recognition}}, pages 891--898, June 2014.
\newblock ISSN: 1063-6919.

\bibitem[\protect\citeauthoryear{Rusu \bgroup \em et al.\egroup
  }{2016}]{rusu_policy_2016}
Andrei~A. Rusu, Sergio~Gomez Colmenarejo, Caglar Gulcehre, Guillaume
  Desjardins, James Kirkpatrick, Razvan Pascanu, Volodymyr Mnih, Koray
  Kavukcuoglu, and Raia Hadsell.
\newblock Policy {Distillation}, January 2016.
\newblock arXiv:1511.06295 [cs].

\bibitem[\protect\citeauthoryear{Shoham \bgroup \em et al.\egroup
  }{2019}]{shoham2019overcoming}
Neta Shoham, Tomer Avidor, Aviv Keren, Nadav Israel, Daniel Benditkis, Liron
  Mor-Yosef, and Itai Zeitak.
\newblock Overcoming forgetting in federated learning on non-iid data.
\newblock {\em arXiv preprint arXiv:1910.07796}, 2019.

\bibitem[\protect\citeauthoryear{Shu \bgroup \em et al.\egroup
  }{2023}]{shu_clustered_2023}
Jiangang Shu, Tingting Yang, Xinying Liao, Farong Chen, Yao Xiao, Kan Yang, and
  Xiaohua Jia.
\newblock Clustered {Federated} {Multitask} {Learning} on {Non}-{IID} {Data}
  {With} {Enhanced} {Privacy}.
\newblock {\em IEEE Internet of Things Journal}, 10(4):3453--3467, February
  2023.
\newblock Conference Name: IEEE Internet of Things Journal.

\bibitem[\protect\citeauthoryear{Silberman \bgroup \em et al.\egroup
  }{2012}]{silberman_indoor_2012}
Nathan Silberman, Derek Hoiem, Pushmeet Kohli, and Rob Fergus.
\newblock Indoor {Segmentation} and {Support} {Inference} from {RGBD} {Images}.
\newblock In Andrew Fitzgibbon, Svetlana Lazebnik, Pietro Perona, Yoichi Sato,
  and Cordelia Schmid, editors, {\em Computer {Vision} – {ECCV} 2012},
  Lecture {Notes} in {Computer} {Science}, pages 746--760, Berlin, Heidelberg,
  2012. Springer.

\bibitem[\protect\citeauthoryear{Smith \bgroup \em et al.\egroup
  }{2017}]{smith2017federated}
Virginia Smith, Chao-Kai Chiang, Maziar Sanjabi, and Ameet~S Talwalkar.
\newblock Federated multi-task learning.
\newblock {\em Advances in neural information processing systems}, 30, 2017.

\bibitem[\protect\citeauthoryear{Thapa \bgroup \em et al.\egroup
  }{2022}]{thapa_splitfed_2022}
Chandra Thapa, Pathum Chamikara~Mahawaga Arachchige, Seyit Camtepe, and Lichao
  Sun.
\newblock {SplitFed}: {When} {Federated} {Learning} {Meets} {Split} {Learning}.
\newblock {\em Proceedings of the AAAI Conference on Artificial Intelligence},
  36(8):8485--8493, June 2022.
\newblock Number: 8.

\bibitem[\protect\citeauthoryear{Wei \bgroup \em et al.\egroup
  }{2021}]{wei_learn_2021}
Qijie Wei, Xirong Li, Weihong Yu, Xiao Zhang, Yongpeng Zhang, Bojie Hu, Bin Mo,
  Di~Gong, Ning Chen, Dayong Ding, and Youxin Chen.
\newblock Learn to {Segment} {Retinal} {Lesions} and {Beyond}.
\newblock In {\em 2020 25th {International} {Conference} on {Pattern}
  {Recognition} ({ICPR})}, pages 7403--7410, January 2021.
\newblock ISSN: 1051-4651.

\bibitem[\protect\citeauthoryear{Wu \bgroup \em et al.\egroup
  }{2020}]{wu2020fedhome}
Qiong Wu, Xu~Chen, Zhi Zhou, and Junshan Zhang.
\newblock Fedhome: Cloud-edge based personalized federated learning for in-home
  health monitoring.
\newblock {\em IEEE Transactions on Mobile Computing}, 21(8):2818--2832, 2020.

\bibitem[\protect\citeauthoryear{Wu \bgroup \em et al.\egroup
  }{2023}]{wu2023fedict}
Zhiyuan Wu, Sheng Sun, Yuwei Wang, Min Liu, Xuefeng Jiang, and Bo~Gao.
\newblock Fedict: Federated multi-task distillation for multi-access edge
  computing.
\newblock {\em arXiv preprint arXiv:2301.00389}, 2023.

\bibitem[\protect\citeauthoryear{Xu \bgroup \em et al.\egroup
  }{2023}]{xu2023demt}
Yangyang Xu, Yibo Yang, and Lefei Zhang.
\newblock Demt: Deformable mixer transformer for multi-task learning of dense
  prediction.
\newblock {\em arXiv e-prints}, pages arXiv--2301, 2023.

\bibitem[\protect\citeauthoryear{Yao and Sun}{2020}]{yao2020continual}
Xin Yao and Lifeng Sun.
\newblock Continual local training for better initialization of federated
  models.
\newblock In {\em 2020 IEEE International Conference on Image Processing
  (ICIP)}, pages 1736--1740. IEEE, 2020.

\bibitem[\protect\citeauthoryear{Yu \bgroup \em et al.\egroup
  }{2020}]{yu2020gradient}
Tianhe Yu, Saurabh Kumar, Abhishek Gupta, Sergey Levine, Karol Hausman, and
  Chelsea Finn.
\newblock Gradient surgery for multi-task learning.
\newblock {\em Advances in Neural Information Processing Systems},
  33:5824--5836, 2020.

\bibitem[\protect\citeauthoryear{Zamir \bgroup \em et al.\egroup
  }{2018}]{zamir_taskonomy_2018}
Amir~R. Zamir, Alexander Sax, William Shen, Leonidas~J. Guibas, Jitendra Malik,
  and Silvio Savarese.
\newblock Taskonomy: {Disentangling} {Task} {Transfer} {Learning}.
\newblock In {\em Proceedings of the {IEEE} {Conference} on {Computer} {Vision}
  and {Pattern} {Recognition}}, pages 3712--3722, 2018.

\bibitem[\protect\citeauthoryear{Zantedeschi \bgroup \em et al.\egroup
  }{2020}]{zantedeschi2020fully}
Valentina Zantedeschi, Aur{\'e}lien Bellet, and Marc Tommasi.
\newblock Fully decentralized joint learning of personalized models and
  collaboration graphs.
\newblock In {\em International Conference on Artificial Intelligence and
  Statistics}, pages 864--874. PMLR, 2020.

\bibitem[\protect\citeauthoryear{Zhang \bgroup \em et al.\egroup
  }{2022}]{zhang2022personalized}
Xu~Zhang, Yinchuan Li, Wenpeng Li, Kaiyang Guo, and Yunfeng Shao.
\newblock Personalized federated learning via variational bayesian inference.
\newblock In {\em International Conference on Machine Learning}, pages
  26293--26310. PMLR, 2022.

\end{thebibliography}
% \nocite{chen_fedmsplit_2022}
% \nocite{ma_over--air_2022}
% \nocite{zhang_federated_2022}
% \nocite{yi_hfedmtl_2022}
% \nocite{cao_resource_2022}
% \nocite{chen_federated_2020}
% \nocite{singh_federated_2021}
% \nocite{xiao_clustered_2021}
% \nocite{li_network_2021}
% \nocite{yu_learning_2020}
% \nocite{he_spreadgnn_2022}
% \nocite{zeng_multi-task_2021}
% \nocite{li_coalition_2023}
\end{document}